\documentclass{bmvc2k}

\usepackage{url}
\usepackage{graphicx}
\usepackage{amsmath}
\usepackage{amssymb}
\usepackage{mathtools}
\usepackage{amsthm}
\usepackage{booktabs}
\usepackage{xcolor}
\usepackage{colortbl}
\usepackage{multirow}
\usepackage[ruled]{algorithm2e}
\usepackage{upgreek}
\usepackage{wrapfig}
\usepackage[symbol]{footmisc}

\title{BDC-Adapter: \\Brownian Distance Covariance for \\ Better Vision-Language Reasoning}

\addauthor{Yi Zhang$^\ast$}{zhangyi2021@mail.sustech.edu.cn}{1,2}
\addauthor{Ce Zhang$^\ast$}{cezhang@cs.cmu.edu}{3}
\addauthor{Zihan Liao}{liaozh2020@mail.sustech.edu.cn}{2}
\addauthor{Yushun Tang}{tangys2022@mail.sustech.edu.cn}{2}
\addauthor{Zhihai He$^\dagger$}{hezh@sustech.edu.cn}{2,4}

\addinstitution{
 Harbin Institute of Technology\\
 Harbin, China
}
\addinstitution{
 Southern University of Science and Technology (SUSTech)\\
 Shenzhen, China
}
\addinstitution{
 Carnegie Mellon University\\
 Pittsburgh, United States
}
\addinstitution{
 Pengcheng Laboratory\\
 Shenzhen, China
}

\runninghead{Zhang \etal}{BDC-Adapter for Better Vision-Language Reasoning}

\def\etal{\emph{et al}\bmvaOneDot}

\begin{document}
\maketitle

\begin{abstract}
Large-scale pre-trained Vision-Language Models (VLMs), such as CLIP and ALIGN, have introduced a new paradigm for learning transferable visual representations. Recently, there has been a surge of interest among researchers in developing lightweight fine-tuning techniques to adapt these models to downstream visual tasks. We recognize that current state-of-the-art fine-tuning methods, such as Tip-Adapter, simply consider the covariance between the query image feature and features of support few-shot training samples, which only captures linear relations and potentially instigates a deceptive perception of independence. To address this issue, in this work, we innovatively introduce Brownian Distance Covariance (BDC) to the field of vision-language reasoning. The BDC metric can model all possible relations, providing a robust metric for measuring feature dependence. Based on this, we present a novel method called BDC-Adapter, which integrates BDC prototype similarity reasoning and multi-modal reasoning network prediction to perform classification tasks. Our extensive experimental results show that the proposed BDC-Adapter can freely handle non-linear relations and fully characterize independence, outperforming the current state-of-the-art methods by large margins. 
\end{abstract}

\section{Introduction}
\label{sec:intro}
Recently, large-scale pre-trained Vision-Language Models (VLMs), such as CLIP \cite{radford2021learning} and ALIGN \cite{jia2021scaling}, have introduced a new paradigm for generic visual recognition \cite{hu2022pushing}. These VLMs jointly learn both visual and textual representations in a shared feature space through pre-training on large-scale datasets retrieved from the Internet, enabling them to recognize a wide range of visual concepts without the need for additional annotated data \cite{ganin2015unsupervised,radford2021learning}.

\begin{figure*}[t]
\centering
\includegraphics[width=\textwidth]{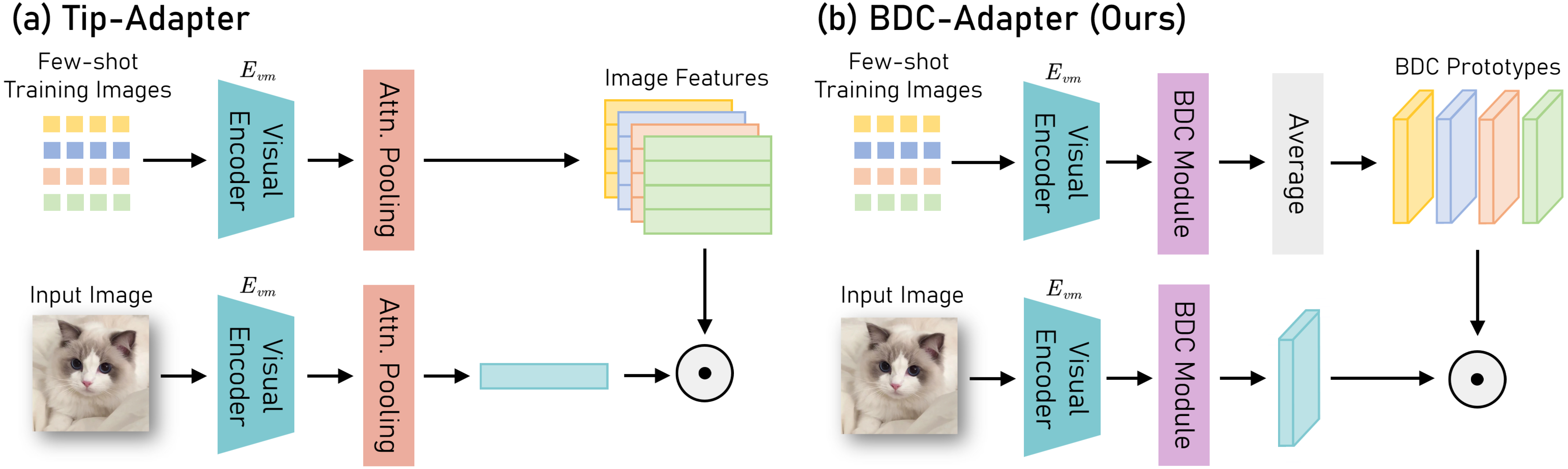}
\vspace{-20pt}
\caption{\textbf{A comparison on Tip-Adapter (left) vs. our proposed BDC-Adapter (right)}. Our BDC-Adapter represents each image by a BDC matrix, which considers the joint distributions and measures non-linear dependence during inference. Note that in this figure,  $E_{vm}$ is the modified image encoder of CLIP without the last attention pooling layer.}
\label{fig:teasor}
\vspace{-18pt}
\end{figure*}

However, due to the massive size of VLMs, it is impractical for individuals to re-train those models. Therefore, lightweight fine-tuning techniques have become essential for adapting VLMs to downstream visual tasks, such as image classification \cite{radford2021learning,zhou2022learning}, object detection \cite{shi2022proposalclip,du2022learning}, and image captioning \cite{li2021grounded,yao2021cpt,wang2022efficient}. One research direction focuses on the \textbf{prompt tuning method}, which aims to learn the prompt from downstream data. For instance, CoOp \cite{zhou2022learning} firstly introduces the prompt tuning method to fine-tune CLIP, while CoCoOp \cite{zhou2022conditional} uses prompts conditioned on model inputs to address the generalization problem. Another approach is the \textbf{adapter-based method}, which directly adapts the extracted features. CLIP-Adapter \cite{gao2021clip} and Tip-Adapter \cite{zhang2022tip} are examples of this approach, both of which introduce feature adapters to enhance CLIP's performance on various downstream tasks.

We notice that the current state-of-the-art Tip-Adapter \cite{zhang2022tip} method, as shown in Figure \ref{fig:teasor}, establishes a key-value cache model and evaluates the similarities of the query image feature and features of support few-shot training samples to perform classification. However, we recognize that  Tip-Adapter \cite{zhang2022tip} simply considers the covariance between each image feature pair, which only measures marginal distributions and captures linear relations. If the relation between features is non-linear \cite{Szekely2007,Szekely2009}, the covariance might be zero, potentially instigating a deceptive perception of independence.
This problem, if not effectively addressed, will hinder our capabilities to fine-tune VLMs.


In this paper, we introduce Brownian Distance Covariance (BDC) to the field of vision-language reasoning to provide a robust metric for measuring feature dependence. 
While classical covariance can only capture linear relations, Brownian covariance can model all possible relations \cite{Szekely2007,Szekely2009}. 
Based on this, we propose a novel approach called BDC-Adapter that leverages BDC to enhance vision-language reasoning ability. 
During the training stage, we first train a one-layer multi-modal reasoning network that learns from few-shot examples across different modalities (\textit{i.e.}, vision and language). Then, we introduce a BDC module that takes feature maps as input and outputs a BDC matrix as a visual representation. Using this, we compute class-specific prototypes by averaging the BDC matrices of the few-shot image samples for each class, which act as a support set for test image classification. In Figure \ref{fig:teasor}, we show the BDC prototype similarity reasoning process of our proposed BDC-Adapter. During the inference stage, we combine the BDC prototype similarity reasoning and multi-modal reasoning network prediction to perform classification tasks.
To evaluate the effectiveness of our BDC-Adapter, we conduct experiments on few-shot learning, domain generalization, and visual reasoning tasks. Our extensive experimental results show that BDC-Adapter outperforms the current state-of-the-art methods by large margins.

\section{Related Work}
\label{sec:related}

{\noindent \bf Fine-Tuning Vision-Language Models.}
In recent studies on VLMs, researchers have explored the semantic correspondence between the textual and visual modalities by leveraging a huge amount of image-text pairs \cite{jia2021scaling,radford2021learning,yu2022coca,wang2022simvlm,cui2022contrastive,yuan2021florence,shukor2022efficient,wang2022image,chen2022pali,geng2023hiclip}. Researchers have demonstrated that with sufficient fine-tuning, the large-scale pre-trained VLMs can be transferred to various downstream tasks, such as image retrieval \cite{lu2019vilbert,duan2022multi}, visual grounding \cite{li2021grounded,yao2021cpt}, semantic segmentation \cite{ma2022open}, and visual question answering \cite{zhou2022unsupervised,duan2022multi,lei2021less}. 

Recent advances in fine-tuning VLMs can be classified into two major categories: \textit{prompt tuning methods} and \textit{adapter-based methods}.
As the pioneering work in the context of prompt tuning, CoOp \cite{zhou2022learning} learns a set of additional vectors to optimize the prompt context. Further, Zhou \etal \cite{zhou2022conditional} extend CoOp to generate image-conditioned vectors to tackle the generalization problem.  TPT \cite{shu2022tpt} can adaptively learn prompts for each test sample in the inference stage. Adapter-based methods directly adapt the extracted visual and textual representations. For example, CLIP-Adapter \cite{gao2021clip} introduces a feature adapter that generates the adapted features to enhance the performance of few-shot recognition. Further, Tip-Adapter \cite{zhang2022tip} proposes a training-free scheme, which achieves higher accuracy by establishing a key-value cache model. UP-Adapter \cite{zhang2023unsupervised} proposes to generate pseudo-labels for the unannotated images, which will be used to train a prototype adapter module.
\\[-8pt]

{\noindent \bf Cross-Modal Few-Shot Image Classification.}
Few-shot learning is an important problem in machine learning, which attempts to enable models' transferability to new tasks with limited labeled examples \cite{tommasi2009more,wang2020generalizing}. Traditional few-shot learning methods typically rely on training from base classes in the source domain, which limits their generalization capabilities to the novel target domains \cite{finn2017model,qi2018low,ye2020few,afham2021rich,zhuo2022tgdm}. 
With the help of large-scale pre-trained VLMs, an alternative direction of work focus on tackling the few-shot classification task without source-domain training \cite{radford2021learning, lin2023multimodality}. By freezing the pre-trained weights and training additional sets of learnable parameters for downstream tasks, these models can achieve remarkable performance with very limited training samples \cite{radford2021learning,zhang2023unsupervised,zhang2023cross}.
\\[-8pt]

{\noindent \bf Brownian Distance Covariance.} 
The BDC metric, defined as the Euclidean distance between the joint characteristic function and the product of the marginals, was first proposed in Sz{\'{e}}kely \etal \cite{Szekely2007,Szekely2009}. While classical covariance can only model linear relations, Brownian covariance can model all possible relations. Therefore, the BDC metric has been introduced into appearance matching \cite{bak2014brownian} and people recognition \cite{bilinski2014representing,bkak2016exploiting} to provide more complementary information for the network model. In recent years, the BDC metric has also been applied in other computer vision applications, such as object detection \cite{wu2023group}, hyperspectral image classification \cite{zhang2021superpixel}, and few-shot learning \cite{xie2022joint} tasks. In this work, we use the BDC metric for representation learning mainly in the few-shot classification setting.

\section{Method}
\begin{figure}[!th]
\centering
\includegraphics[width=\textwidth]{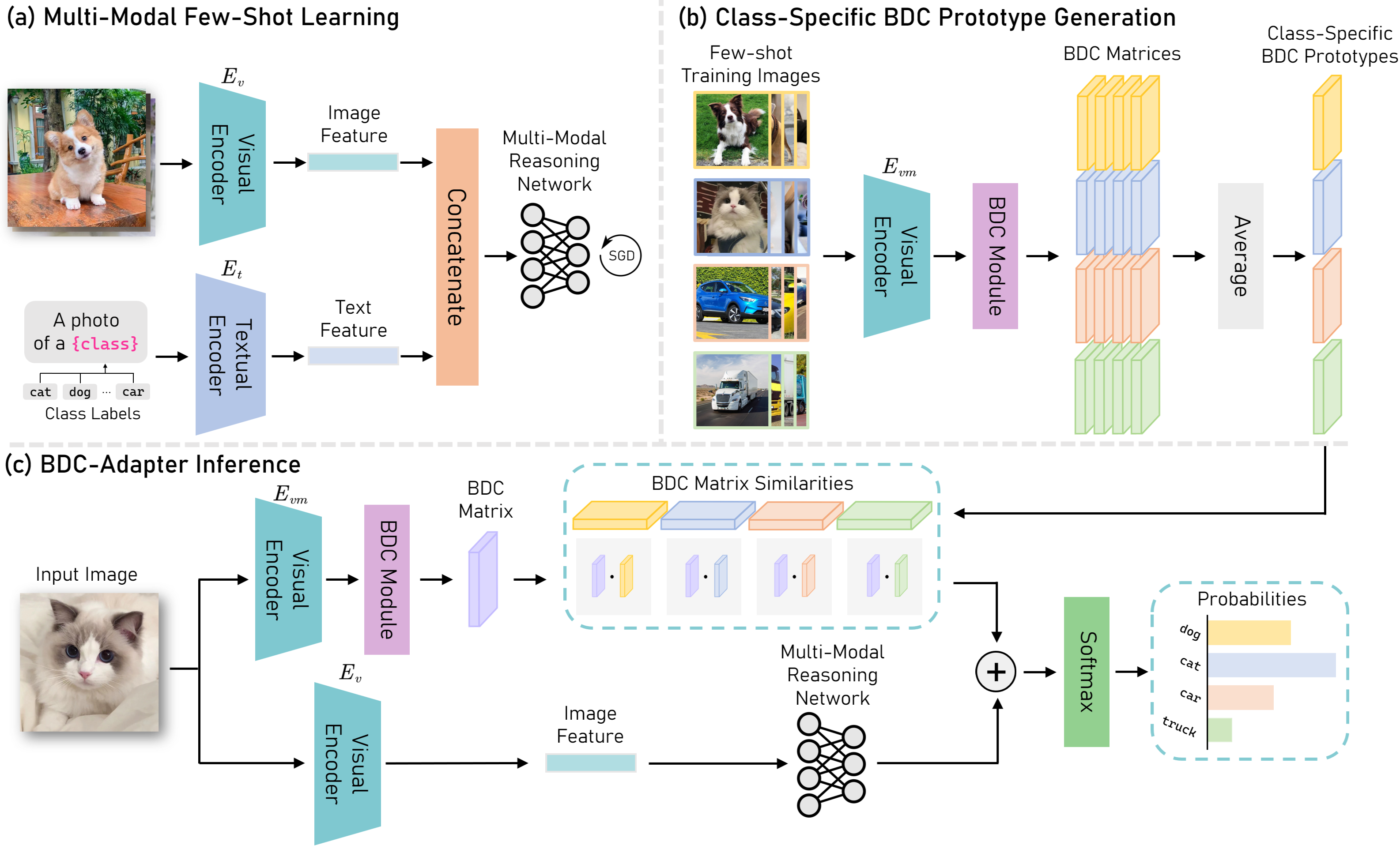}
\vspace{-22pt}
\caption{\textbf{Overview of our BDC-Adapter method}. $E_v$ and $E_t$ are the original image and text encoders of CLIP respectively, and $E_{vm}$ is the modified image encoder of CLIP that does not include the last attention pooling layer. \textbf{(a)} shows the multi-modal few-shot learning process, \textbf{(b)} shows the class-specific BDC prototype generation process, and \textbf{(c)} presents the whole BDC-Adapter inference process.}
\label{fig:overview}
\vspace{-15pt}
\end{figure}
\subsection{Background}
{\noindent \bf Contrastive Language-Image Pre-Training}. CLIP \cite{radford2021learning} has demonstrated remarkable performance on visual tasks by encoding images and text descriptions onto a shared embedding space and exploiting contrastive learning on noisy image-text pairs on the Internet.
We denote CLIP's encoders as $\{E_t, E_v\}$, where $E_t$ is the text encoder (typically a Transformer \cite{vaswani2017attention}), $E_v$ is the image encoder (typically a ResNet \cite{he2016deep} or ViT \cite{dosovitskiy2020image}). 
In the zero-shot scenario, given a test image $x_{test}$ for a $N$-class classification problem, we utilize CLIP's encoders to extract the visual feature $f_{v} = E_v(x_{test})$ and $N$ text features $f_{t_i} = E_t(\{\pi; y_i\})$ for all classes, where the class name $y_i$ is appended to a hand-crafted prompt $\pi$, such as ``a photo of a''.
The prediction probability on $x_{test}$ can be computed as
\begin{equation}
\label{eq-clip}
   p(y = y_i|x_{test})=\frac{\exp \left( \mathrm{sim}\left(f_{t_i},f_{v} \right) /\tau \right)}{\sum\nolimits_{t'}{\exp \left( \mathrm{sim}\left( f_{t'},f_{v} \right) /\tau \right)}}, 
\end{equation}
where $\tau$ is the temperature hyper-parameter of the softmax function, and $\mathrm{sim}(\cdot, \cdot)$ indicates the cosine similarity.
\\[-8pt]

{\noindent \bf Brownian Distance Covariance}.
The concept of Brownian Distance Covariance (BDC) was first formalized in the literature by Sz{\'{e}}kely \etal~\cite{Szekely2007,Szekely2009}, with a foundation in characteristic function theory.
Suppose $X \in \mathbb{R}^{p}, Y \in \mathbb{R}^{q}$ denote random vectors with dimensions $p$ and $q$ respectively, and $p_{XY}(\mathbf{x},\mathbf{y})$ represents their joint probability density function (PDF). With $\mathbf{t}$ standing as the characteristic of the distribution $X$ and $\mathbf{s}$ for that of $Y$, the joint characteristic functions of $(X,Y)$, expressed as $f_{XY}(\mathbf{t},\mathbf{s})=\mathcal{F}(p_{XY}(\mathbf{x},\mathbf{y}))$, embody a collection of functions that encapsulate the interrelation within the random distributions $X$ and $Y$. Here, $\mathcal{F}$ is a mapping from the distribution space to the feature space. The selection of $f_{XY}(\mathbf{t},\mathbf{s})$ can be diverse, and in our experiments, we employ a network as the characteristic function.
In accordance with the definitions of the joint and marginal characteristic functions, and assuming that random vectors $X$ and $Y$ possess finite first moments, the BDC metric, quantifying the similarity between the characteristics of distributions $X$ and $Y$, can be expressed as
\begin{align}\label{equ:definition BDC}
\vspace{-3pt}
\mathcal{V}(X,Y)=\int_{\mathbf{t}\in\mathbb{R}^{p}}\int_{\mathbf{s}\in\mathbb{R}^{q}}\dfrac{\Vert f_{XY}(\mathbf{t},\mathbf{s})-f_{X}(\mathbf{t})f_{Y}(\mathbf{s})\Vert^2}{c_{p}c_{q}\Vert\mathbf{t}\Vert^{1+p}\Vert\mathbf{s}\Vert^{1+q}}d\mathbf{s}d\mathbf{t}.
\vspace{-3pt}
\end{align}
Here, $\Vert\cdot\Vert$ denotes the Euclidean norm, $c_{p}=\pi^{(1+p)/2}/\Gamma((1+p)/2)$, and $\Gamma$ represents the complete gamma function. Despite Equation~(\ref{equ:definition BDC}) is complicated in its current form, the BDC metric possesses a closed-form expression for discrete observations as established in the work by Sz{\'{e}}kely \etal \cite{Szekely2009}, which is elaborated in the following Section \ref{subsection:BDC-computation}.

\subsection{BDC Module for Image Representation}\label{subsection:BDC-computation}

Let the pair of observation data matrices $(\mathbf{T}, \mathbf{S})$ represent the observation of the joint characteristic function $f_{XY}(\mathbf{t},\mathbf{s})$. Here, the $i$-th observation's $\mathbf{t}_i,\mathbf{s}_i$ constitute the $i$-th column of $\mathbf{T}$ and $\mathbf{S}$ respectively.
The Euclidean distance matrix, derived from each observational pair of $X$, is denoted as $\mathbf{D}^{\mathbf{T}}_1=(d^{\mathbf{T}}_{k,l})\in \mathbb{R}^{m\times m}$, where $d^{\mathbf{T}}_{k,l}=\Vert\mathbf{t}_{k}-\mathbf{t}_{l}\Vert$.

The Euclidean distance matrix $\mathbf{D}^{\mathbf{T}}_1$ can be calculated by first computing the squared Euclidean distance matrix $\mathbf{D}^{\mathbf{T}}_2$ and subsequently taking the square root. Through this process, we can obtain  a closed-form expression of $\mathbf{D}^{\mathbf{T}}_1$ in terms of $\mathbf{T}$:

\begin{equation}\label{equ:computaton of D}
\mathbf{D}^{\mathbf{T}}_1=\sqrt{\mathbf{D}^{\mathbf{T}}_2},\quad
\mathbf{D}^{\mathbf{T}}_2=\mathbf{1}(\mathbf{T}^{\top}\mathbf{T}\circ \mathbf{I})+(\mathbf{T}\mathbf{T}^{\top}\circ \mathbf{I})\mathbf{1}-2\mathbf{T}^{\top}\mathbf{T}.
\end{equation}
In this equation, $\mathbf{1}\in \mathbb{R}^{d\times d}$ is a matrix where each element equals $1$, $\mathbf{I}$ signifies the identity matrix, $\circ$ denotes the Hadamard product, and $\top$ represents the matrix transpose.

The entry located at the $k$-th row and $l$-th column $r^{\mathbf{T}}_{k,l}$ of the so-called \textit{BDC matrix}, denoted as $\mathbf{R}^{\mathbf{T}}=(r^{\mathbf{T}}_{k,l})$, is defined relative to the Euclidean distance $d^{\mathbf{T}}_{k,l}$ as
\begin{equation}
   r^{\mathbf{T}}_{k,l}=d^{\mathbf{T}}_{k,l}-\frac{1}{m}\sum_{k=1}^{m}d^{\mathbf{T}}_{k,l}-\frac{1}{m}\sum_{l=1}^{m}d^{\mathbf{T}}_{k,l}-\frac{1}{m^{2}}\sum_{k=1}^{m}\sum_{l=1}^{m}d^{\mathbf{T}}_{k,l},
\end{equation}
where the final three terms represent the means of the $l$-th column, $k$-th row, and all entries of $\mathbf{D}^{\mathbf{T}}_1$, respectively. Consequently, we can represent $\mathbf{R}^{\mathbf{T}}$ concerning $\mathbf{D}^{\mathbf{T}}_1$ as
\begin{equation}\label{equ:computaton of R}
\mathbf{R}^{\mathbf{T}}=\mathbf{D}^{\mathbf{T}}_1-\frac{1}{d}\big(\mathbf{1}\mathbf{D}^{\mathbf{T}}_1+{\mathbf{D}^{\mathbf{T}}_1}^{\top}\mathbf{1}\big)+\frac{1}{d^{2}}\mathbf{1}\mathbf{D}^{\mathbf{T}}_1\mathbf{1}.
\end{equation}
The matrix $\mathbf{R}^{\mathbf{S}}$ can be derived analogously from $\mathbf{S}$. Subsequently, the BDC metric assumes the ensuing form, as per \cite{Szekely2009}:
\begin{align}\label{equ: BDC measure}
\mathcal{V}(X,Y)=\mathrm{tr}\big({\mathbf{R}^{\mathbf{T}}}^{\top}\mathbf{R}^{\mathbf{S}}\big).
\end{align}
Here, $\mathrm{tr}(\cdot)$ denotes the matrix trace.

Referring to the above derivation, it is clear that the BDC metric facilitates an explicit expression in terms of the feature matrix. Subsequent to this, the construction of the BDC module proceeds as follows. Specifically, we devise a dual-layer module, which firstly reduces feature dimension and then computes the BDC matrix. \textbf{(1) Dimension Reduction:} Owing to the polynomial increase in the computation complexity of the BDC matrix with respect to the number of channels of the feature, we incorporate a convolution layer for the purpose of dimension reduction.
\textbf{(2) Calculation of BDC Matrix:} Assuming that the previous layer embeds the input image $\mathbf{u}\in \mathbb{R}^{H\times W\times 3}$ into the feature represented by a $g \times d$ tensor and each column or each row of the tensor can be considered as a characteristic of observation from $X$. In the second layer, we calculate the BDC matrix according to  Equation~(\ref{equ:computaton of D}) and Equation~(\ref{equ:computaton of R}). It should be noted that this layer contains no learnable parameters.

As a result, we characterize the BDC module as a training-free pooling layer. Deriving from Equation~(\ref{equ:computaton of D}) and (\ref{equ:computaton of R}), it is  obvious that the BDC matrix encapsulates non-linear interrelations among channels via the Euclidean distance.  Consequently, when the relation between features is non-linear, the traditional covariance might be zero, which may potentially instigate a deceptive perception of independence. In contrast, the BDC is invariably non-negative and only amounts to zero when the features are indeed independent~\cite{Szekely2007,Szekely2009}. This constitutes an advantage over conventional covariance, thereby positioning BDC as a more robust metric for evaluating dependence between features.

\subsection{BDC-Adapter for CLIP}
\label{sec:BDC_for_CLIP}
An overview of our proposed BDC-Adapter method is shown in Figure \ref{fig:overview}. In this section, we introduce each component of our proposed BDC-Adapter method in detail. 
\\[-8pt]

{\noindent \bf Multi-Modal Few-Shot Learning.}
Following prior works \cite{lin2023multimodality}, we first construct text samples by appending the class label $y_i$ to a hand-crafted prompt such as $\pi =$ ``a photo of a", then we get the text descriptions $t_i = \{\pi; y_i\}$ for each class $y_i$ in all $N$ classes. In each training batch, we randomly sample $n_1$ text descriptions $\{t_i\}_{i=1}^{n_1}$, $n_2$ image samples $\{x_i\}_{i=1}^{n_2}$ and their labels $\{y_i\}_{i=1}^{n}$, where $n = n_1 + n_2$ is the number of samples in a batch. We then extract the text feature or image feature $f_i$ for each sample, denoted as $f_i = E_v(x_i)$ for images or $f_i = E_t(t_i)$ for texts. Here we use $f_i$ for both features since the text or image samples will be projected onto the same dimensional embedding space by encoders of CLIP. Note that the multi-modal features $\{f_i\}_{i=1}^{n}$ in each batch is also L2 normalized. Based on these features, we learn a one-layer multi-modal reasoning network $\psi$ to classify the image, which can be denoted as
 \begin{equation}
\label{eq:m_feature}
    \psi(x) = W^{\top}x,
\end{equation}
where $W$ is the parameter of the multi-modal reasoning network $\psi$ initialized with text features by $w_{y_i}=E_t(t_i), \forall i \in [1, N]$, where $w_{y_i}$ is the classification weight for class $y_i$ in parameter matrix $W$.
The weights in this linear layer can be updated by gradient descent with the following cross-entropy loss during training:
\begin{equation}
\label{eq:celoss}
    \mathcal{L} _{CE}=\sum_{i=1}^n{H\left( y_i,\psi \left( f_i \right) \right)}=-\sum_{i=1}^n{\log \left( \frac{e^{w_{y_i}\cdot f_i}}{\sum\nolimits_{y'}{e^{w_{y'}\cdot f_i}}} \right)}.
\end{equation}

{\noindent \bf Class-Specific BDC Prototype Generation.}
\label{sec:generation}
In a few-shot learning task, it provides $M$-shot $N$-class training samples (\textit{i.e.} $M$ annotated images in each of the $N$ categories) in a new dataset. We can denote the $M$ images in a class as $\{x_m\}_{m=1}^M$ and the class labels as $\{y_n\}_{n=1}^N$.  For each image $x$ within class $y$, we first utilize a modified visual encoder of CLIP $E_{vm}$ to generate its L2 normalized visual feature, then feed it to a BDC module to produce a BDC matrix $B_y(x)$.  Given all the BDC matrix $\{B_y(x_m)\}_{m=1}^M$ of $M$ images within class $y$, we define the prototype of class $y$ to be the average of the BDC matrices, denoted as 
${P}_{y}=\frac{1}{M}\sum_{m=1}^M{B}_y({x}_{m})$. 
Therefore, for the entire training dataset, we can build a prototype set 
$\mathcal{P} = \{P_{y_n}\}_{n=1}^N$. 
\\[-8pt]

{\noindent \bf BDC-Adapter Inference.}
During inference, for a test image $x_{test}$, we first utilize the visual encoder to extract its image feature $f_{test} =  E_v(x_{test})$. 
Therefore, the prediction of the multi-modal reasoning network can be denoted as
\begin{equation}
\label{eq-P_m}
    p_m(y=y_n|x_{test})= w_{y_{n}} \cdot f_{test},
\end{equation}
where $1\leqslant n\leqslant N$ is the class index.

After the BDC prototype generation process, we have obtained the prototype of the few-shot training samples. Similarly, we utilize $E_{vm}$ to generate the image feature of $x_{test}$, then feed it into the BDC module and obtain the BDC matrix $B\left(x_{test}\right)$. 
We can get the prediction of image $x_{test}$ via calculating the similarity between $B\left(x_{test}\right)$ and the prototypes in the set $\mathcal{P}$, denoted as
\begin{equation}
\label{eq-P_b}
    p_b(y=y_n|x_{test})= \exp \left( -\delta \left(1-\mathrm{vec}\left(B\left(x_{test}\right)\right) \cdot \mathrm{vec}\left(P_{y_n}\right)\right) \right),
\end{equation}
where $\mathrm{vec}(\cdot)$ denotes the vectorization of a matrix. The term $\mathrm{vec}\left(B\left(x_{test}\right)\right) \cdot \mathrm{vec}\left(P_{y_n}\right)$ is equivalent to the cosine similarities between the BDC matrix of the test image $x_{test}$ and the prototype matrix $P_{y_n}$. The exponential function is adopted to convert the similarities into non-negative values with $\delta$ adjusting its sharpness. 

We then combine $p_m$ and $p_b$ to get the final prediction, 
\begin{equation}
\begin{aligned}
\label{eq-LOGIT}
     {p(y=y_n|x_{test})} &\ = \alpha p_b(y=y_n|x_{test})+ p_m(y=y_n|x_{test})  \\ 
              &\ = \alpha \exp \left( -\delta \left(1-\mathrm{vec}\left(B\left(x_{test}\right)\right) \cdot \mathrm{vec}\left(P_{y_n}\right)\right) \right)  + w_{y_{n}} \cdot f_{test},
\end{aligned}
\end{equation}
where $\alpha$ is the residual ratio to combine two predictions. Note that $w_{y_{n}}$ is the weight for class $y_n$ in the linear layer $\psi$ and can be updated by $\mathcal{L}_{CE}$ defined in Equation (\ref{eq:celoss}) during training. The final predicted label of test image $x_{test}$ is produced by 
$\hat{y} = \underset{y'}{\mathrm{arg}\max}\,p(y'|x_{test})$.

For clarity, we also analyze the sensitivity levels of the hyper-parameters and provide the pseudo-code of our method in the Supplemental Materials.

\section{Experiment}

\subsection{Experiment Setup} \label{sec:setup}
To comprehensively evaluate the performance of our proposed BDC-Adapter method, we conduct experiments on few-shot image recognition, domain generalization, and visual reasoning tasks.
For \textbf{few-shot image recognition}, we follow prior methods~\cite{zhou2022learning,zhang2022tip} and adopt the common few-shot protocol to evaluate our method on 11 well-known image classification datasets, including generic object classification, fine-grained object classification, remote sensing recognition, texture classification, scene recognition, and action recognition:  ImageNet~\cite{recht2019imagenet}, Caltech101~\cite{fei2004learning},  OxfordPets~\cite{parkhi2012cats}, StandfordCars~\cite{krause20133d}, Flowers102~\cite{nilsback2008automated}, Food-101~\cite{bossard2014food}, FGVC Aircraft~\cite{maji2013fine},  DTD~\cite{cimpoi2014describing}, SUN397~\cite{xiao2010sun}, EuroSAT~\cite{helber2019eurosat}, and UCF101~\cite{soomro2012ucf101}. These datasets provide a comprehensive benchmark to evaluate the few-shot learning performance of each method.
For \textbf{domain generalization}, we evaluate the  model’s robustness to natural distribution shifts by training on 16-shot ImageNet~\cite{deng2009imagenet} and testing on four variants of ImageNet: ImageNet-V2~\cite{recht2019imagenet}, ImageNet-Sketch~\cite{wang2019learning}, ImageNet-A~\cite{hendrycks2021natural}, and ImageNet-R~\cite{hendrycks2021many}. Those variant datasets have been treated as out-of-distribution data for ImageNet in previous work \cite{zhou2022conditional,shu2022tpt}.
For \textbf{visual reasoning on human object interaction (HOI)}, we conduct experiments on Bongard-HOI~\cite{jiang2022bongard} benchmark to evaluate the effectiveness of our proposed BDC-Adapter method on visual reasoning tasks.

\subsection{Implementation Details} \label{sec:implementation}
Our BDC-Adapter is based on CLIP \cite{radford2021learning} with ResNet-50 image encoder and Transformer text encoder. In the training stage, we freeze the weights to inherit the prior knowledge.  Note that $E_{vm}$ defined in Section \ref{sec:BDC_for_CLIP} is the modified image encoder of CLIP that does not include the last attention pooling layer, whose output will be fed to the BDC module. For multi-modal few-shot learning introduced in Section \ref{sec:BDC_for_CLIP}, there are no requirements for the image and text samples to be exactly matched, and the number of samples for each modality can vary in a batch, which means $n_1$ (the number of image samples) is not always equal to $n_2$ (the number of text samples). Therein, $n_1$ is equal to the number of shots for training (\textit{i.e.}, 1, 2, 4, 8, 16 in our experiments).  Following prior methods, we apply the data pre-processing protocol in CLIP \cite{radford2021learning}, such as resizing and random cropping operations, \textit{etc}. On all the datasets in our experiments, we train our model for 30 epochs and set the initial learning rate as $1\times 10^{-3}$. The AdamW~\cite{kingma2015adam} optimizer with a cosine annealing scheduler is used to optimize the parameters. Our method is parameter-efficient and lightweight, and we only use one single NVIDIA RTX 3090 GPU for training. 
For the visual reasoning on HOI task, we further introduce the implementation details in the Supplemental Materials.

\subsection{Performance Analysis} \label{sec:Performance Analysis}
\begin{figure*}[t]
\centering
\includegraphics[width=\textwidth]{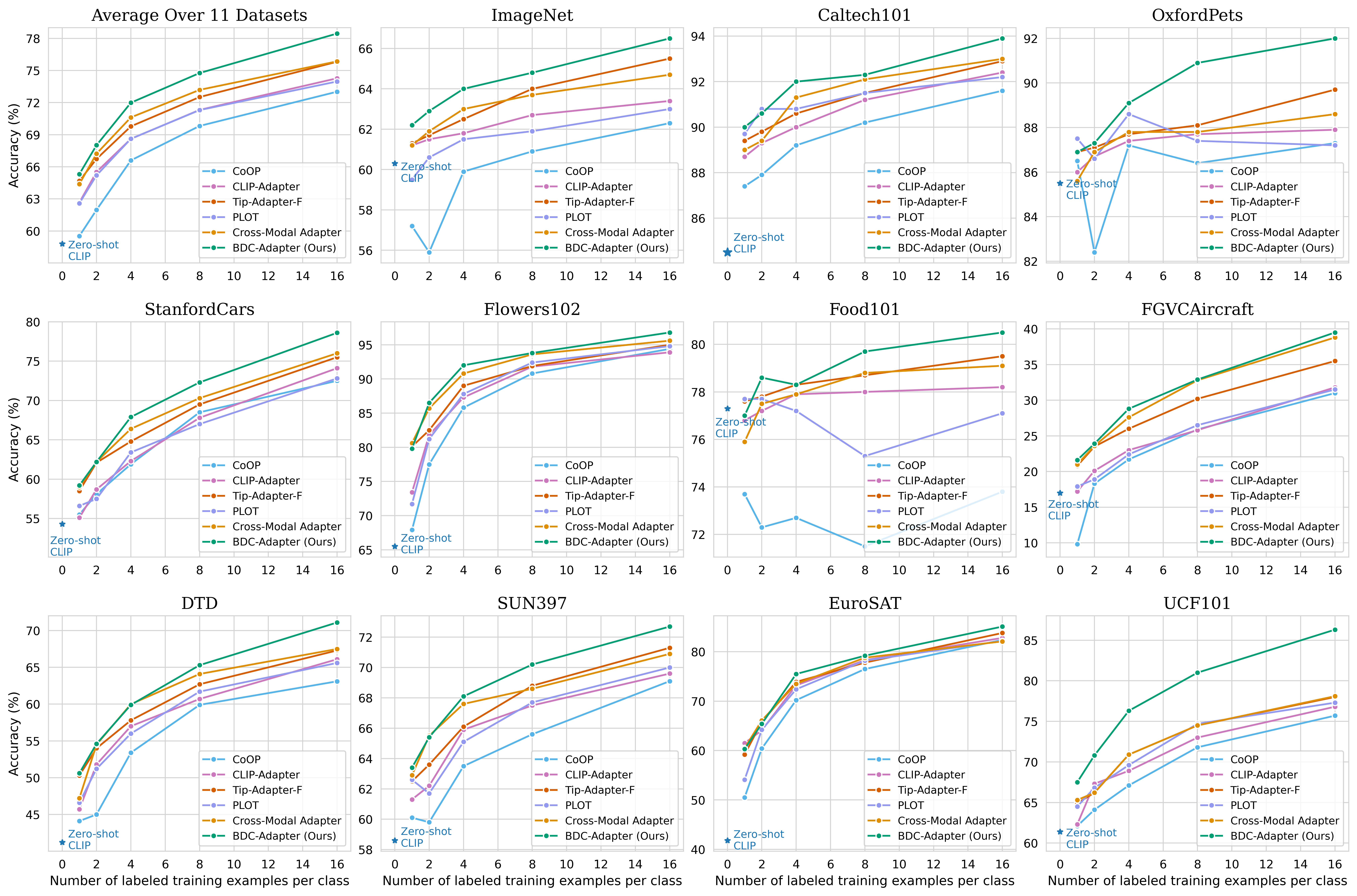}
\vspace{-20pt}
\caption{\textbf{Performance comparisons on few-shot learning on 11 datasets}. For each dataset, we report the accuracy on 1-/2-/4-/8-/16-shot settings. The top-left subfigure shows the average accuracy over all 11 datasets.}
\label{fig:fewshot_results}
\vspace{-5pt}
\end{figure*}

\subsubsection{Few-Shot Learning}
Figure \ref{fig:fewshot_results} compares the performance of our method with five baseline methods on all 11 datasets. We also present the average accuracy in the top-left sub-figure of Figure \ref{fig:fewshot_results}. We observe that our BDC-Adapter outperforms other state-of-the-art methods and obtains the highest average accuracy.
In comparison to Tip-Adapter-F \cite{zhang2022tip} (a fine-tuned version of Tip-Adapter), our method consistently outperforms it by large margins on all 11 datasets. This proves that our proposed BDC-Adapter can capture the non-linear relations ignored by Tip-Adapter-F \cite{zhang2022tip} and fully characterize independence.

We notice that our method experiences a performance drop  when using 4 shots on Food101, which appears to be a common overfitting challenge encountered not only by our approach but also by several existing adaptation methods like PLOT~\cite{chen2023plot} and CoOp~\cite{zhou2022learning}. 
However, the overall results have demonstrated the effectiveness of our BDC-Adapter.

\begin{wraptable}[12]{r}{0.6\textwidth}
\small
\begin{center}
\vspace{-40pt}
\caption{\textbf{Performance comparisons on robustness to natural distribution shifts}. All the experiments are conducted with ResNet-50 visual backbone. The best results are in \textbf{bold} and the second are \underline{underlined}.}
\vspace{-5pt}
\label{table:generalization}
\resizebox{0.6\textwidth}{!}{
\begin{tabular}{lcccccc}
\toprule
\multirow{2}*{Method}  & Source & \multicolumn{5}{c}{Target} \\ \cmidrule(lr){2-2} \cmidrule(lr){3-7}  & ImageNet & -V2 & -Sketch & -A & -R  & Avg. \\
\midrule
Zero-Shot CLIP~\cite{radford2021learning}  &  60.33 &  53.27  & 35.44  & 21.65 &  56.00  & 41.59\\
Linear Probe CLIP~\cite{radford2021learning}  & 56.13  & 45.61  & 19.13  & 12.74  & 34.86 &  28.09\\
CoOp~\cite{zhou2022learning}   & 63.33  & 55.40  & 34.67  & 23.06  & 56.60 &  42.43\\
CoCoOp~\cite{zhou2022conditional} &   62.81  & 55.72  & 34.48  & 23.32  & 57.74  & 42.82\\
ProGrad~\cite{zhu2022prompt} &   62.17  & 54.70  & 34.40  & 23.05  & 56.77  & 42.23\\
PLOT~\cite{chen2023plot} &   63.01  & 55.11  & 33.00  & 21.86  & 55.61  & 41.40\\
DeFo~\cite{wang2023learning} &    64.00  & \textbf{58.41}  & 33.18  & 21.68  & 55.84  & 42.28\\
TPT~\cite{shu2022tpt} &   60.74 &  54.70  & 35.09  & 26.67  & \underline{59.11}  & 43.89\\
TPT + CoOp~\cite{shu2022tpt} &   \underline{64.73}  & 57.83  & \underline{35.86}  & \underline{30.32}  & 58.99  & \underline{45.75}\\
\rowcolor{gray!20}
\textbf{BDC-Adapter (Ours)} &   \textbf{66.46}  & \underline{58.05}  & \textbf{36.92}  & \textbf{30.77}  & \textbf{59.52}  & \textbf{46.31}\\
\bottomrule
\end{tabular}
}
\end{center}
\end{wraptable}

\subsubsection{Domain Generalization}
Table \ref{table:generalization} summarizes the performance of our proposed BDC-Adapter and other state-of-the-art methods. For a fair comparison, we directly include the results of other baselines from their original paper. We report the classification accuracy of the source domain (ImageNet \cite{deng2009imagenet}), target domain (ImageNet-V2~\cite{recht2019imagenet}, ImageNet-Sketch~\cite{wang2019learning}, ImageNet-A~\cite{hendrycks2021natural}, and ImageNet-R~\cite{hendrycks2021many}), and the average accuracy on out-of-distribution data. Our method outperforms the DeFo~\cite{wang2023learning} method on 3 out of 4 target datasets, and surpasses all other baselines in all metrics. These results indicate that our BDC-Adapter exhibits remarkable robustness to distribution shifts.

\subsubsection{Visual Reasoning on Bongard-HOI} 
\begin{figure*}[b]
\vspace{-14pt}
\centering
\includegraphics[width=\textwidth]{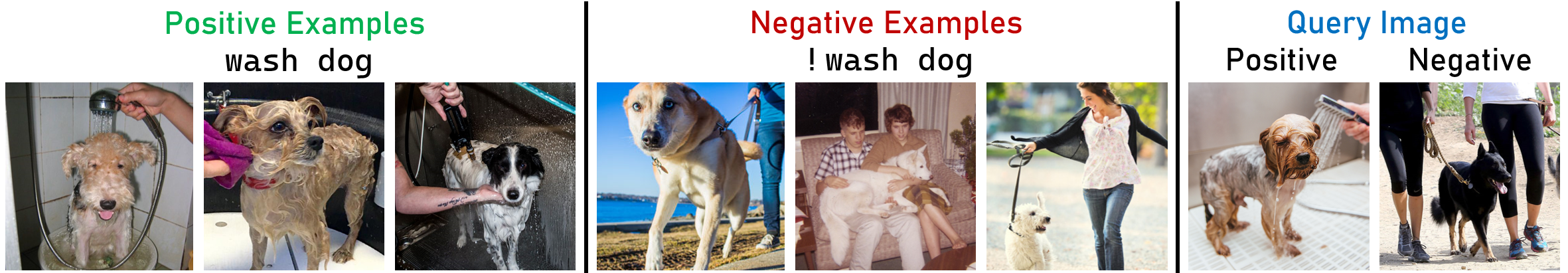}
\vspace{-16pt}
\caption{\textbf{Illustration of a few-shot learning instance from the
Bongard-HOI \cite{jiang2022bongard} benchmark}.  The left side shows positive images that depict the visual relationship of a person washing a dog. In contrast, negative examples do not exhibit such relationships. The right side shows query images, where the ground-truth labels are positive or negative, respectively.}
\label{fig:hoi_example}
\end{figure*}
\begin{wraptable}[10]{r}{0.64\textwidth}
\vspace{-10pt}
\caption{\textbf{Performance comparisons  on the Bongard-HOI \cite{jiang2022bongard} dataset}. The last column shows the average accuracy. The best results are in \textbf{bold}  and the second are \underline{underlined}.}
\vspace{2pt}
\label{table:hoi}
\centering
\resizebox{0.62\textwidth}{!}{
\begin{tabular}{lccccc}
\toprule
\multirow{3}{*}{Method} & \multicolumn{5}{c}{Test Splits}          \\
\cmidrule(lr){2-6}
                        & Seen act.     & Unseen act.     & Seen act.     & Unseen act.     & \multirow{2}{*}{Avg.}    \\
                        & Seen obj.     & Seen obj.     & Unseen obj.     & Unseen obj.     &     \\
\midrule
CNN-Baseline \cite{nie2020bongard}                   & 50.03 & 49.89 & 49.77 & 50.01 & 49.92 \\
Meta-Baseline \cite{chen2020new}                   & 58.82 & 58.75 & 58.56 & 57.04 & 58.30 \\
ProtoNet  \cite{snell2017prototypical}                  & 58.90 & 58.77 & 57.11 & 58.34 & 58.28 \\
HOITrans \cite{zou2021end}                    & 59.50 & 64.38 & 63.10 & 62.87 & 62.46 \\
TPT (RN50) \cite{shu2022tpt}                    & \underline{66.39} & \underline{68.50} & \underline{65.98} & \underline{65.48} & \underline{66.59} \\
\rowcolor{gray!20}
\textbf{BDC-Adapter} (RN50)    & \textbf{68.36} & \textbf{69.15} & \textbf{67.67} & \textbf{67.82} & \textbf{68.25} \\
\bottomrule
\end{tabular}
}
\end{wraptable}
In Figure \ref{fig:hoi_example}, we illustrate some instances in the Bongard-HOI dataset \cite{jiang2022bongard}. Note that there are 6 positive examples, 6 negative examples, and 1 query image in a test instance, which is different from the illustration here. 
Following the experimental design outlined in Jiang \etal \cite{jiang2022bongard}, the comparison is conducted on four distinct test splits of the Bongard-HOI dataset. 
It should be noted that the results for the other baselines are sourced directly from the research paper by Jiang \etal \cite{jiang2022bongard}. For more details on this task, interested readers are directed to this paper.
We compare the performance of the proposed BDC-Adapter approach with previous approaches in Table \ref{table:hoi}. 
Remarkably, our method outperforms the conventional methods, including ProtoNet \cite{snell2017prototypical} and HOITrans \cite{zou2021end}, by large margins. Even compared to the CLIP-based TPT method, BDC-Adapter still yields better performance in all 4 test splits.

\begin{wraptable}[7]{r}{0.5\textwidth}
\vspace{-34pt}
\caption{\textbf{Effectiveness of different components in our BDC-Adapter method}. MRN represents multi-modal reasoning network and BDC represents BDC prototype similarity reasoning, init. stands for initialization.}
\vspace{2pt}
\label{table:ablation}
\centering
\resizebox{\linewidth}{!}{
\begin{tabular}{lccccc}
\toprule
Few-shot Setup & 1    & 2  & 4  & 8 & 16 \\ \midrule
MRN (\textit{w/o} init.)& 60.55 & 61.07 & 61.89 & 63.04 & 63.57 \\
MRN (\textit{w/} init.)         & 61.12 & 61.77 & 62.73 & 63.78 & 64.68 \\
\rowcolor{gray!20}
\textbf{MRN + BDC (Ours)}             & \textbf{62.19} & \textbf{62.91} & \textbf{63.95} & \textbf{64.83} &\textbf{66.46} \\
\bottomrule
\end{tabular}
}
\end{wraptable}

\subsection{Ablation Study}
To systematically evaluate the effectiveness of our proposed BDC-Adapter, we conduct an ablation study on the ImageNet \cite{deng2009imagenet} dataset to analyze the impacts of different components in our BDC-Adapter. Table \ref{table:ablation} presents the performance results, where the last row shows the accuracy of our full BDC-Adapter. We can see that both initialization of the multi-modal reasoning network and BDC prototype similarity reasoning contribute significantly to the overall performance.

\subsection{Efficiency Comparison}
\begin{wraptable}[7]{r}{0.5\textwidth}
\centering
\vspace{-15pt}
\caption{\textbf{Efficiency comparisons on 16-shot ImageNet}. We report the results using a single NVIDIA RTX 3090 GPU.}
\vspace{2pt}
\label{table:efficiency}
\resizebox{\linewidth}{!}{
\begin{tabular}{l|cccccc}
\toprule
Method   & Epochs    & Training & GFLOPs  & Param. & Acc. \\
\midrule
CoOp & 200   & 15 h & $>$10 & \textbf{0.01M}   & 62.95 \\
CLIP-Adapter & 200   & 50 min & 0.004 & 0.52M   & 63.59 \\
Tip-Adapter-F & 20   & 5 min & 0.030 & 16.3M   & 65.51 \\
\rowcolor{gray!20}
\textbf{Ours} & \textbf{20}   & \textbf{2 min} & \textbf{0.001} & 1.02M   & \textbf{66.46} \\
\bottomrule
\end{tabular}
}
\end{wraptable}
In order to show the great fine-tuning efficiency of our BDC-Adapter, we compare the number of training epochs, training time, computational cost, and number of parameters of our method with other state-of-the-art methods on 16-shot ImageNet using a single NVIDIA RTX 3090 GPU. We report the comprehensive results in Table \ref{table:efficiency}. Our BDC-Adapter has only a single linear layer for training, thereby exhibiting great efficiency in fine-tuning VLMs.
With just 2 minutes of training and 1 MFLOP on a single RTX 3090, our BDC-Adapter achieves a remarkable accuracy of 66.46\% on 16-shot ImageNet. In comparison, the CoOp method needs about 15 hours of training and 4 MFLOPs to achieve 62.26\% accuracy; the Tip-Adapter-F method needs 5 minutes of training and 30 MFLOPs to achieve  65.51\% accuracy.

\section{Conclusion}
In this work, we innovatively introduce Brownian Distance Covariance to the field of vision-language reasoning, which provides a more robust metric for measuring feature dependence to enable beter generalization capability. Based on this, we present a novel method called BDC-Adapter, which takes advantage of the BDC metric in computing the similarities between the few-shot BDC prototypes and the BDC matrix of the test image. Meanwhile, BDC-Adapter only introduces a one-layer multi-modal reasoning network that learns from multi-modal few-shot instances, to adapt VLMs to downstream tasks using limited training data.
Our extensive experiment results indicate the effectiveness of our proposed BDC-Adapter method for fine-tuning VLMs.
With its lightweight and parameter-efficient design, BDC-Adapter not only exhibits better vision-language reasoning capabilities but also has lower computational complexity, which makes it suitable for practical applications.

\bibliography{egbib}
\end{document}